\documentclass[final,5p,times,twocolumn,authoryear]{elsarticle}
\graphicspath{ {./figures/} }
\usepackage{csquotes}
\usepackage{float}
\usepackage{listings}

\usepackage[T1]{fontenc}
\usepackage{verbatim} %comments
\usepackage{apalike}
\usepackage{graphicx}
\usepackage[figuresright]{rotating}
\usepackage{subcaption}
\usepackage{array}
\usepackage{caption}
\usepackage{tabularx}
\usepackage{amsmath}
\usepackage{float}
\usepackage{booktabs}
\usepackage{geometry}
\usepackage{pdflscape}
\usepackage{afterpage}
\usepackage{rotating}
\restylefloat{figure}
\restylefloat{table}
\usepackage{hyperref}
\usepackage{xcolor} 
\hypersetup{
    colorlinks=true,
    linkcolor=blue,
    citecolor=blue,
    urlcolor=blue,
}
\usepackage{amssymb}
\usepackage{lipsum}

\journal{Expert systems with applications}

\begin{document}

\begin{frontmatter}

\title{DataViz3D: An Novel Method Leveraging Online Holographic Modeling for Extensive Dataset Preprocessing and Visualization }

\author[first]{Jinli Duan}
\affiliation[first]{organization={New York University, Tandon School of Engineering},
            city={New York},
            state={NY},
            country={United States}}
            \ead{jd5374@nyu.edu}

\begin{abstract}
DataViz3D is an innovative online software that transforms complex datasets into interactive 3D spatial models using holographic technology. This tool enables users to generate scatter plot within a 3D space, accurately mapped to the XYZ coordinates of the dataset, providing a vivid and intuitive understanding of the spatial relationships inherent in the data. DataViz3D's user-friendly interface makes advanced 3D modeling and holographic visualization accessible to a wide range of users, fostering new opportunities for collaborative research and education across various disciplines. This project is hosted at:\href{https://cadapp-7f2806a069ea.herokuapp.com/}{DataViz3D Frontpage}
\end{abstract}


\end{frontmatter}

\section{Introduction}

In the area of big data, the ability to effectively visualize and interpret complex datasets is of paramount importance. DataViz3D, a novel online software tool, represents a significant step forward in this domain, leveraging advanced holographic modeling to enhance the process of dataset visualization and preprocessing.This tool offers an intuitive perception of showing spatial relationships inherent within the data using holographic display, a perspective that traditional two-dimensional visualizations often fail to provide.

The human visual system, being the dominant sensory pathway for information processing. Research suggests that a significant portion of the information we process is mediated through vision\cite{visualbrain}.
This underscores the importance of data visualization, defined as the graphical representation of information\cite{Graphicalrep}. DataViz3D meets this need by enabling a dynamic and interactive exploration of data in a three-dimensional space. This approach not only enhances the visualization of three-dimensional data but also offers new perspectives for understanding high-dimensional data, which can be challenging to represent in flat, two-dimensional formats.

The application of holographic technology in DataViz3D is particularly significant. It facilitates a dynamic representation of data in three-dimensional space, enabling users to engage with and comprehend data in ways that two-dimensional visualizations cannot achieve\cite{holographic}. By utilizing holograms, DataViz3D allows for an enhanced visual experience where data points are not only visible but also interactable in a simulated three-dimensional environment. 

DataViz3D’s user-friendly interface further broadens its appeal, making advanced 3D modeling and holographic visualization accessible to a diverse range of users. This accessibility fosters new avenues for collaborative research and education across various disciplines. The software's intuitive design ensures that users, regardless of their technical background, can effectively engage with and analyze complex data structures.

In the following sections, we will dive deeper into the technical aspects of DataViz3D, explore its application structure, and discuss its potential impact on the field of data analysis and visualization. The aim is to demonstrate how DataViz3D stands as a transformative tool, enhancing our ability to interact with and understand complex datasets in the field of big data.
%一个数据点可以通过其在三维空间中的位置来表示三个维度的数据，同时它的颜色深浅可以代表第四维度的数据，球体的大小可以代表第五维度的数据。这样，我们就能在一个三维的视觉空间中同时观察到多个维度的数据，从而更全面地理解数据集的复杂性和内在关系。这种方法在复杂数据集的可视化中非常有用，尤其是在科学研究、金融分析等领域。可以将球体替换成其他几何体来表示高维数据。在数据可视化中，使用不同的几何形状是一种常用的技术，用于表示和区分数据集中的多个维度。例如，除了球体之外，还可以使用立方体、锥体、圆柱体等不同的几何形状。每种几何形状都可以有其独特的属性来表示不同的数据维度，例如：形状自身的维度：如立方体的长度、宽度和高度可以代表三个不同的数据维度。颜色、纹理和透明度：这些视觉属性可以用来表示额外的数据维度。位置和方向：几何体在空间中的位置和朝向也可以作为数据的表现形式。通过这种方式，可以将数据的多个维度映射到一个三维空间中的可视化对象上，从而使得在三维空间内观察和理解多维数据成为可能。这种方法在交互式数据可视化和复杂数据分析中非常有效，尤其是在需要同时展示多个数据维度的情况下

\section{Related Work}

In reviewing the previous work of researches that forms the foundation of our study, this section explores pivotal contributions and technological advancements in data visualization. We focus particularly on the developments that have led to the integration of three-dimensional and holographic visualization techniques, which are central to the design and functionality of DataViz3D.

\subsection{3D-RadViz}
 3D-RadViz\cite{3D-Radviz} extends the traditional Radial Visualization (RadViz) by introducing a third dimension, offering an improved way to visualize complex datasets with minimal overlap and ambiguity. This method has been successfully applied to various datasets, demonstrating its efficacy in providing clear and distinct visual representations, especially in cases where traditional 2D approaches, like RadViz2D, might fall short. The development of 3D-RadViz is a significant contribution to the field of data visualization, addressing the challenges of interpreting multi-dimensional data in a more intuitive and accessible manner.
 
 The development of 3D-RadViz has been inspiring for projects like DataViz3D, particularly in how it addresses the challenge of visualizing high-dimensional data. By adding a third dimension, 3D-RadViz demonstrates a novel way to reduce overlap and enhance clarity in data representation.
\begin{figure}[ht!]
    \centering
    \includegraphics[width=0.75\linewidth]{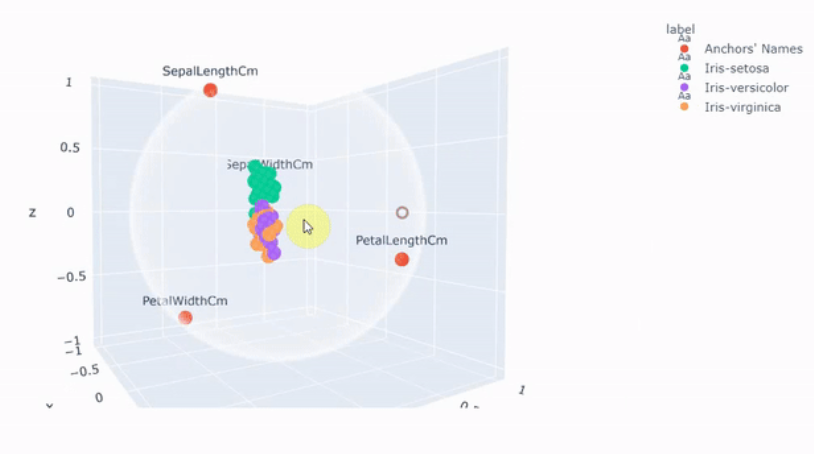}
    \caption{3D-RadViz Iris}
    \label{fig:enter-label}
\end{figure}
\subsection{Parallel Coordinate Plot (PCP)}
Another example is Parallel Coordinate Plot (PCP)\cite{pcp-ed}, a technique for visualizing multidimensional data, allowing the relationships among various dimensions to be explored in a two-dimensional format. Despite its utility, the interpretation of a PCP can be heavily influenced by the ordering of its axes, as adjacent attributes are more readily compared by viewers. Challenges arise when dealing with datasets that have a large number of attributes, often leading to cluttered visualizations that can obscure insights. 
\begin{figure}[ht!]
    \centering
    \includegraphics[width=0.75\linewidth]{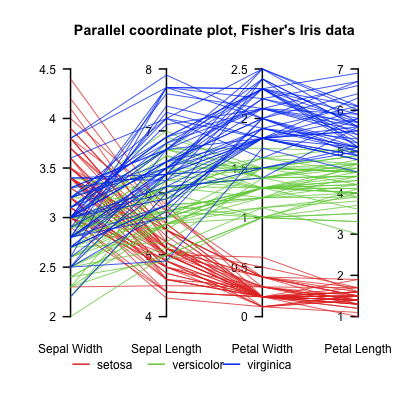}
    \caption{PCP Iris}
    \label{fig:enter-label}
\end{figure}
\subsection{Matplotlib}
Based on the  offical website, Matplotlib has been discribed as a "comprehensive library for creating static, animated, and interactive visualizations in Python"\cite{Matplotlib}.It's widely used for its robustness and ability to produce a wide range of high-quality graphs and plots focusing on 2D, including histograms, power spectra, bar charts, error charts, scatter plots, and more. For 3D visualizations, Matplotlib's mplot3d toolkit extends its capability to plot in three dimensions, allowing for surface, wireframe, scatter, and bar plots in 3D. 

Matplotlib's limitations in 3D and holographic visualization highlighted a gap that DataViz3D aims to fill, offering enhanced interaction and dimensional depth for complex data analysis.
\begin{figure}[ht!]
    \centering
    \includegraphics[width=0.75\linewidth]{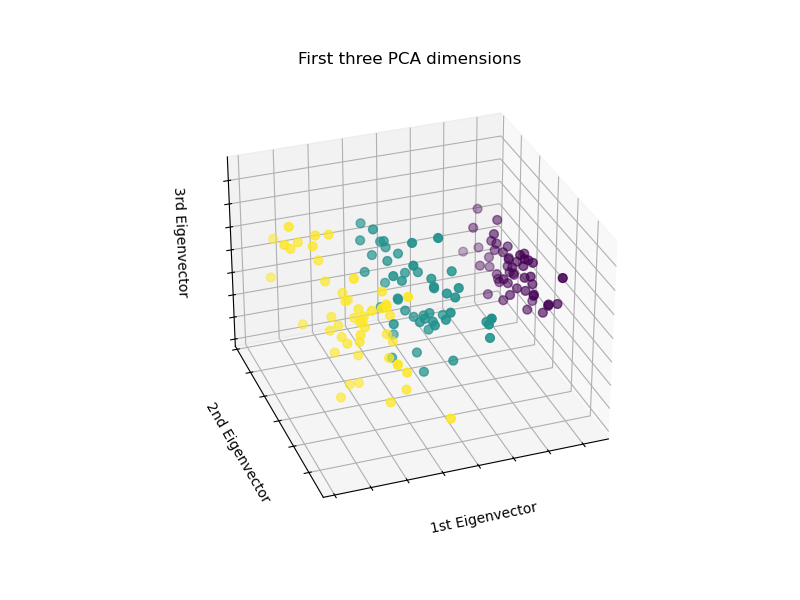}
    \caption{Matplot Iris-3d}
    \label{fig:enter-label}
\end{figure}

\begin{figure*}
  \centering
  \includegraphics[width=0.75\textwidth]{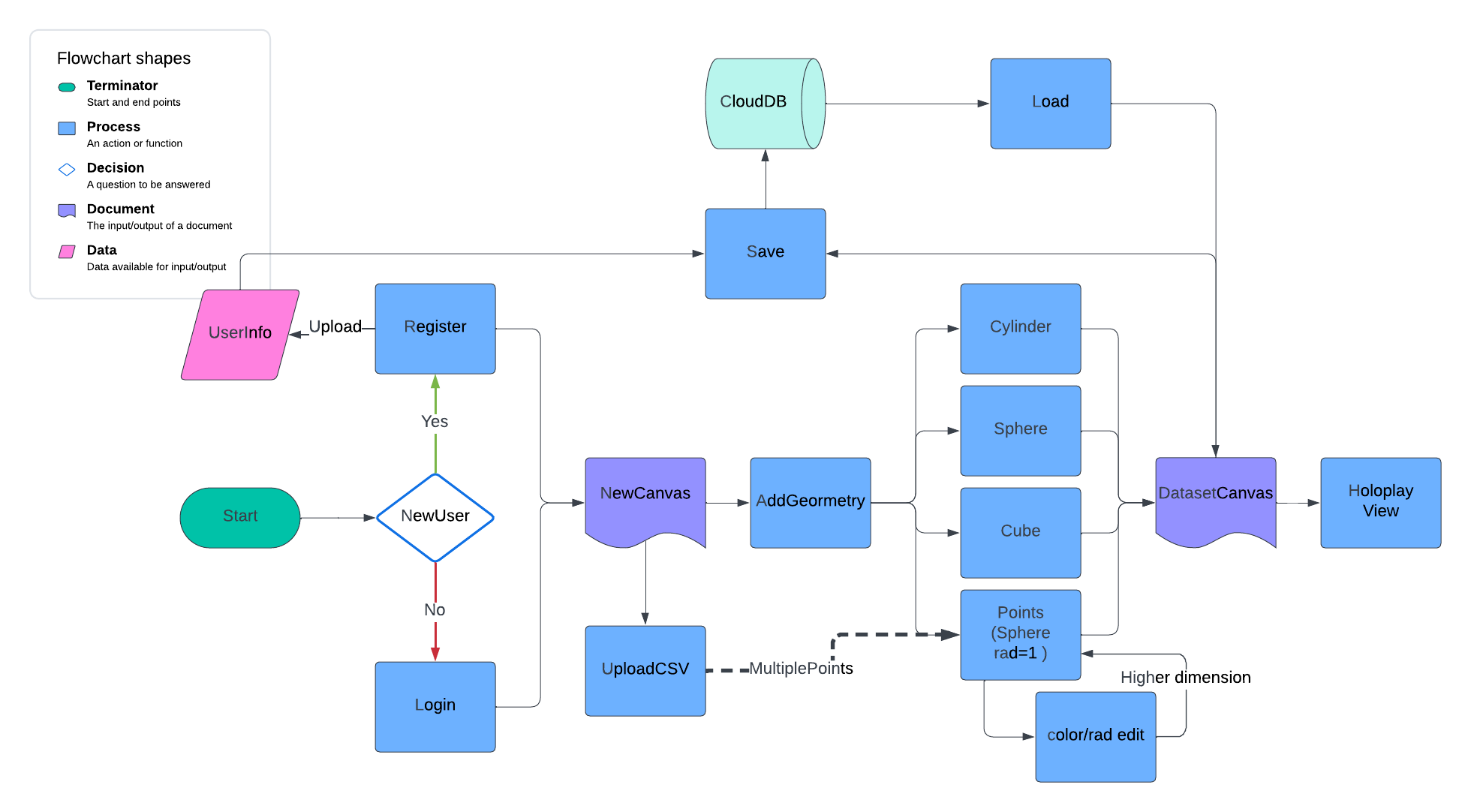}
  \caption{Flowchart-DataViz3D}
  \label{fig:figure1}
\end{figure*}

\section{DataViz3D}

DataViz3D is a comprehensive 3D data visualization tool that allows users to create interactive and complex multi-dimensional representations. It enables the addition of geometric shapes, such as cubes, cylinders, and spheres, to a canvas, where points can be positioned based on XYZ coordinates. These points, customizable in size and color, can represent up to six dimensions of data. Users can also upload CSV files for batch generation of points, streamlining the visualization of large datasets. The interactive canvas supports manipulation and saving of visualizations to user accounts. Further enhancing its capabilities, DataViz3D offers holographic projection on Looking Glass devices, providing an immersive experience in data analysis and presentation.

\subsection{Three.js}
Three.js is one of the most popular JavaScript libraries for rendering 3D graphics in a web browser\cite{threejs}. In our project, Three.js is used to create and manipulate 3D environments and objects, allowing for detailed and interactive visualization of data. The library provides a wide array of features such as creating various geometries, handling lighting, textures, and materials, and controlling the camera and viewpoint. This makes Three.js an essential tool for building immersive 3D visualizations and interactive experiences in the web-based platform of DataViz3D.
\subsection{Looking Glass Factory}
Looking Glass Factory\cite{LookingGlass_2024} is a company known for its development of holographic display technology. Their products are designed for displaying holographic 3D content, offering a unique medium for visualizing data and images in a way that appears to float in physical space. It creates a more immersive and interactive experience, as users can view and interact with 3D visualizations without the need for VR/AR headsets. 
\subsubsection{Holoplay.js}
Holoplay.js\cite{holoplay}, developed by Looking Glass Factory, employs a unique mechanism for displaying holograms. It creates up to 100 distinct views of a 3D scene, presented over a wide view cone. This technique enables the perception of 3D objects through two key methods: parallax, where changing the viewer's position alters the scene's aspect, and stereo vision, offering different perspectives to each eye. The technology uses a "quilt" of images - a collection of different views - which Holoplay.js and related software transform into holographic displays.
\begin{figure}[ht!]
    \centering
    \includegraphics[width=0.75\linewidth]{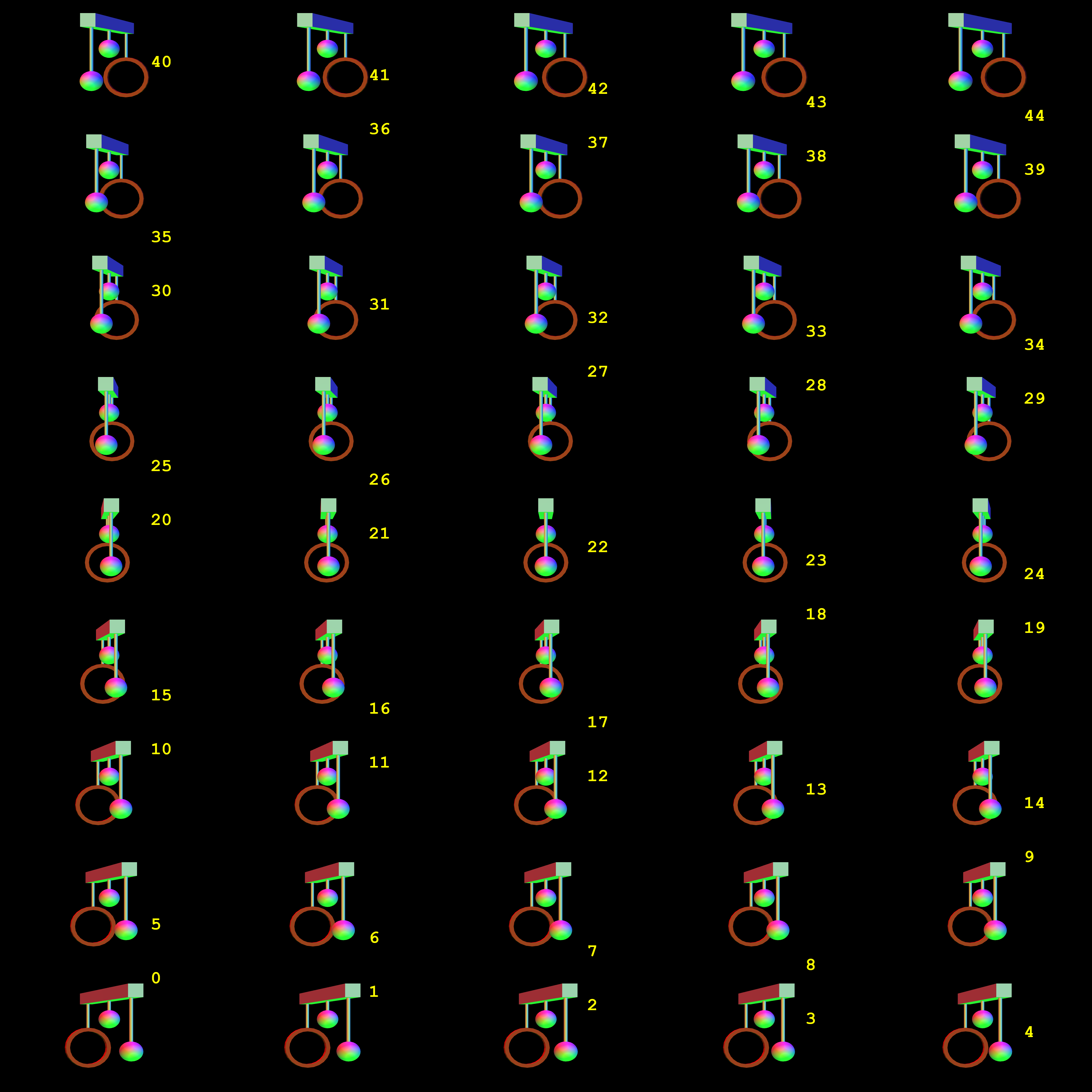}
    \caption{Holoplay mechanism}
    \label{fig:enter-label}
\end{figure}
\subsection{3d Objects}
This aspect of DataViz3D involves creating and manipulating 3D objects such as cubes, spheres, and cylinders within the visualization environment. These objects serve as visual representations of data points, offering users the ability to customize their appearance — including size, color, and position — to effectively represent different data dimensions. This capability allows for a more dynamic and detailed exploration of data in three-dimensional space.
\subsection{Camera\&renderer}

In DataViz3D, camera, rendering, and lighting are key features based on WebGL. Three.js\cite{threejs} enables adjustments in camera angles, zoom, and rotation for diverse views. The renderer renders and displays these settings and 3D objects, ensuring clarity and interactivity. Lighting is also added depth and realism to the visualization.
\subsection{High-dimensional display}
In this system, a single data point can represent up to three dimensions based on its position in the 3D space. Additional dimensions are depicted through variable attributes of the geometric shapes, such as the depth of color indicating a fourth dimension, and the size of a shape, like a sphere, representing a fifth dimension. Users can also switch between different geometric shapes, like cubes or cylinders, to represent more dimensions, with each shape's unique attributes like length, width, and texture conveying different data aspects. This multi-dimensional visualization technique is particularly valuable in areas like scientific research and financial analysis, where understanding intricate data relationships is crucial. The high degree of user customization in representing these dimensions makes DataViz3D a powerful tool for data analysis.

In the Iris example given below was inspired by \cite{Irismethod}: the first three columns by default are set to determine the X, Y, and Z coordinates. The fourth dimension is depicted through the size of the shape, such as the radius of a sphere, which is determined by the value in the fourth column. The fifth dimension is represented by the color of the data point, with the fifth column specifying the color value. Additional columns are not accounted for in the current visualization but can be retained for potential future enhancements. 

\section{Experiments}
In the experiment section, we demonstrate the application of DataViz3D using the well-known Iris dataset\cite{misc_iris_53}, which comprises five dimensions: sepal length, sepal width, petal length, petal width, and variety. This dataset provides an ideal scenario for showcasing the software's capabilities in visualizing multi-dimensional data. In our setup, the sepal length, sepal width, and petal length serve as the XYZ coordinates for each data point in the 3D space. The petal width is used to determine the radius of each point, effectively representing the fourth dimension. The variety dimension, encompassing three categories, is visually distinguished through color-coding—red, yellow, and blue—providing an intuitive method for classifying the data points.

The screenshots from DataViz3D’s canvas illustrate the interactive 3D plot of the Iris dataset, where the positioning, size, and color of each point are in direct relation to the dataset's attributes. The holographic projection, achieved through Holoplay.js, uses a technique involving slicing of images to create a holographic effect. This method allows for a deeply immersive visualization experience, demonstrating the potential of DataViz3D in presenting data in a more engaged manner. The holographic images effectively showcase how DataViz3D can transform traditional data analysis into a more dynamic and visually striking process.
\begin{figure}[ht!]
    \centering
    \includegraphics[width=0.75\linewidth]{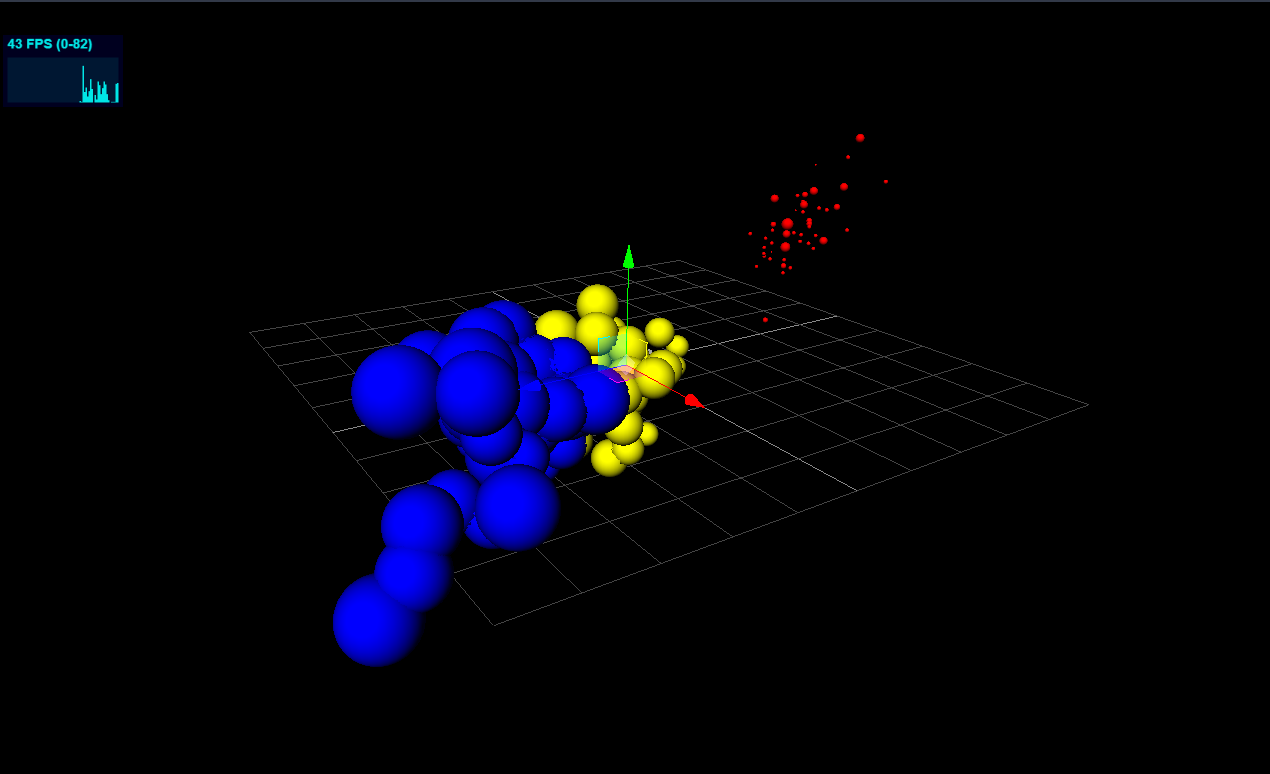}
    \caption{DataViz3D canvas front view}
    \label{fig:enter-label}
\end{figure}
\begin{figure}[ht!]
    \centering
    \includegraphics[width=0.75\linewidth]{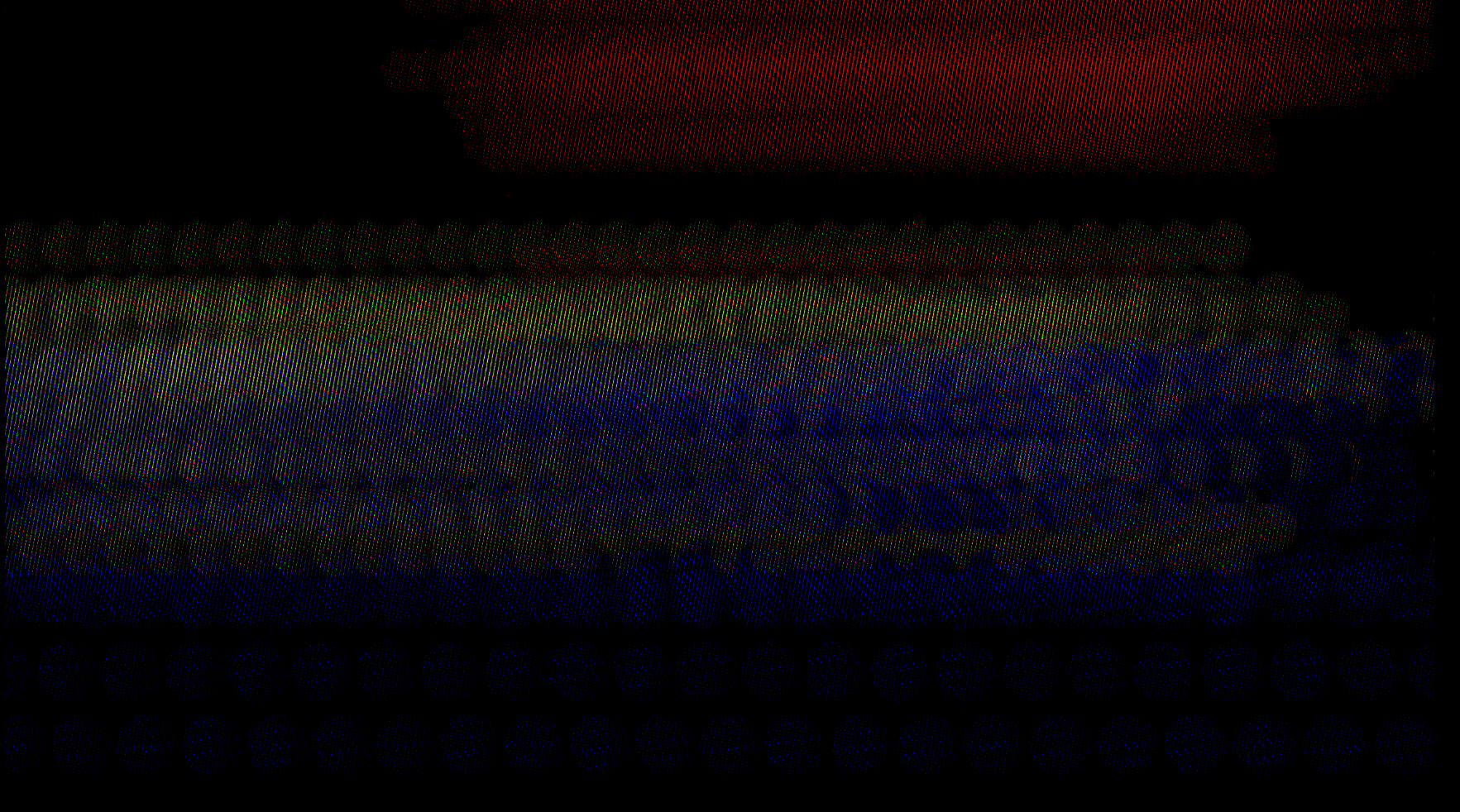}
    \caption{DataViz3D Holoplay front view}
    \label{fig:enter-label}
\end{figure}
\begin{figure}[ht!]
    \centering
    \includegraphics[width=0.75\linewidth]{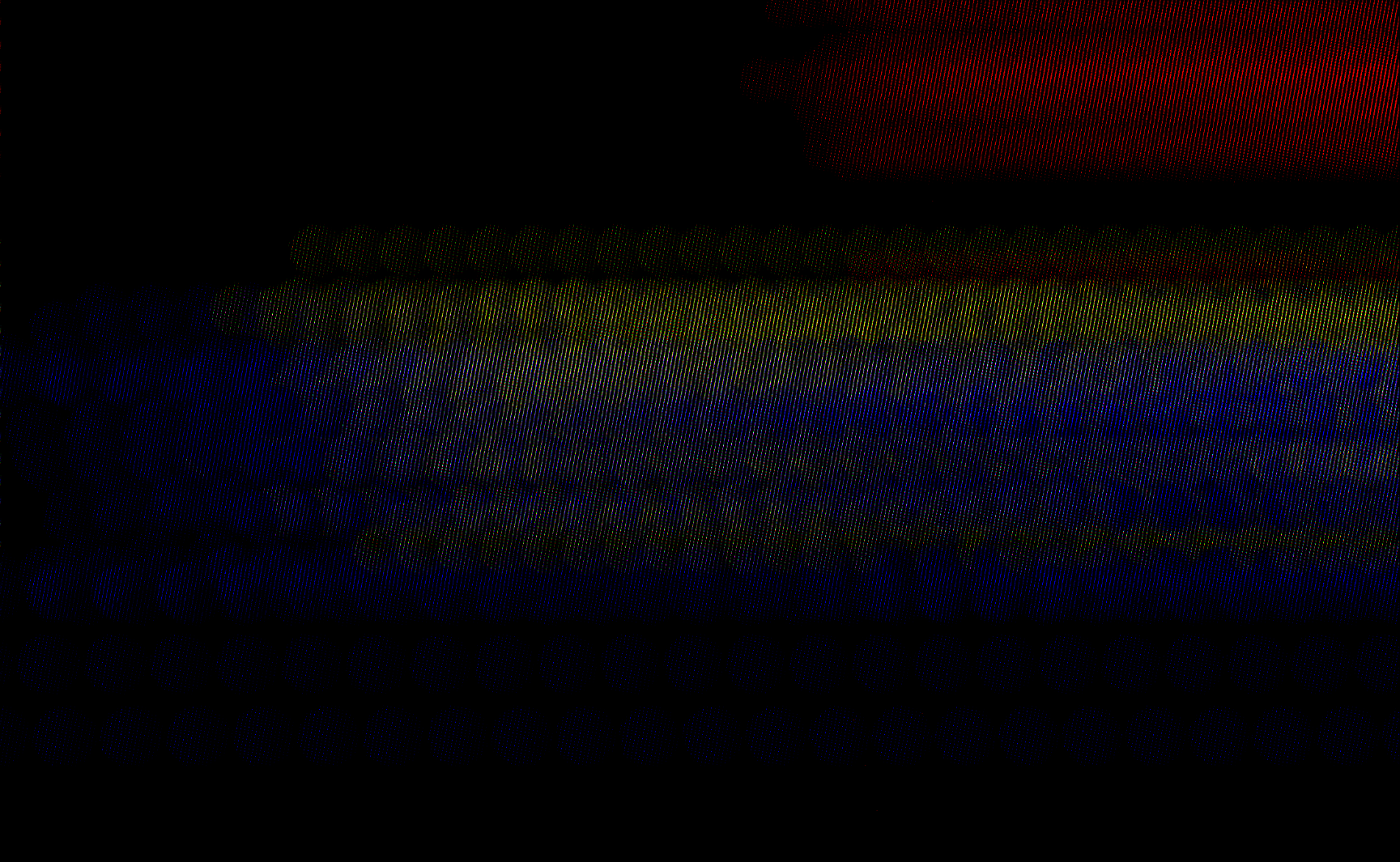}
    \caption{DataViz3D Holoplay left 30 degree view}
    \label{fig:enter-label}
\end{figure}
\begin{figure}[ht!]
    \centering
    \includegraphics[width=0.75\linewidth]{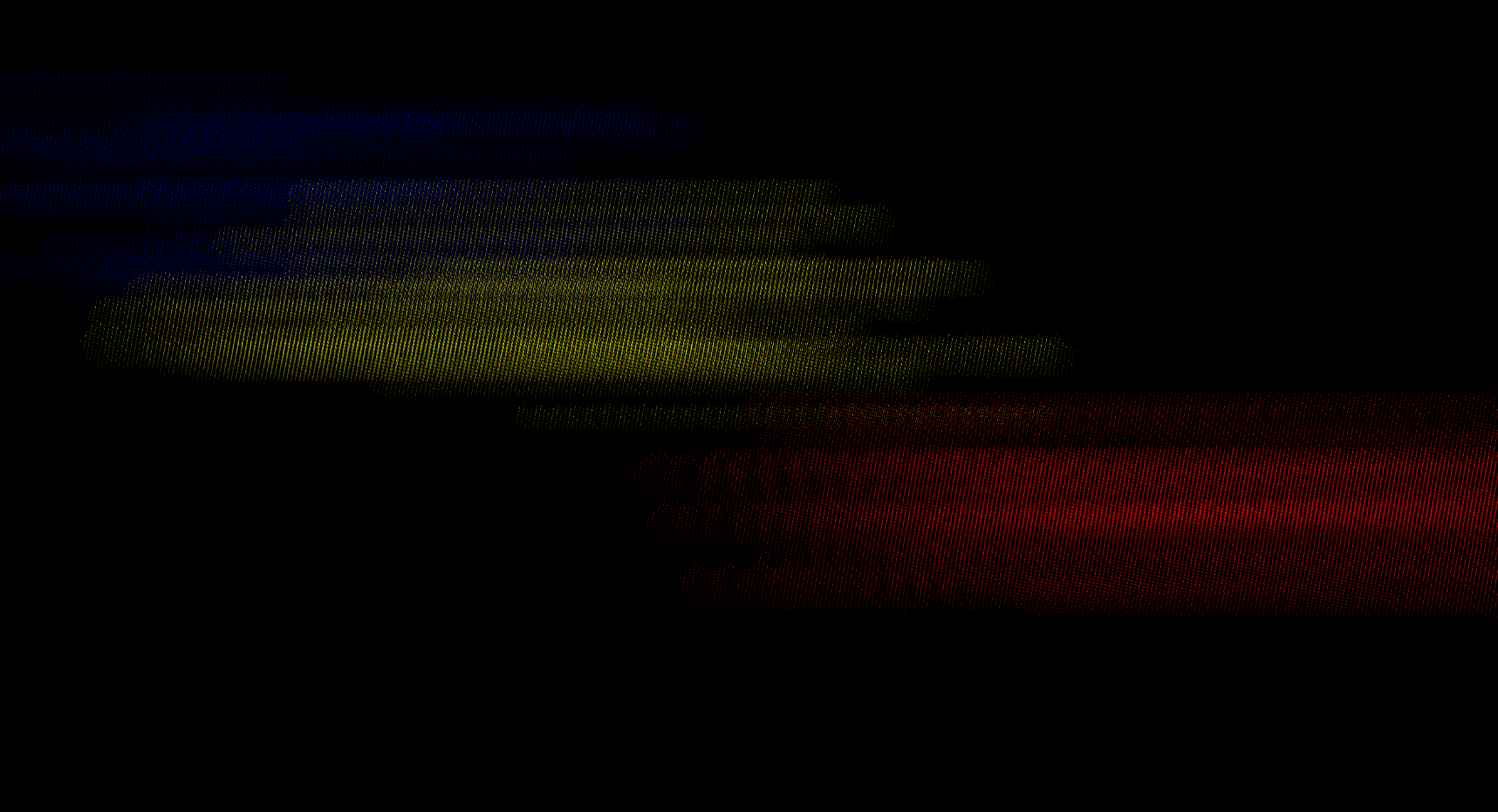}
    \caption{DataViz3D Holoplay right 30 degree view}
    \label{fig:enter-label}
\end{figure}
\section{Conclusion}

DataViz3D, introduces a novel approach to comprehensively preprocess and visualize extensive datasets. This tool not only enhances the interpretation of complex data through intuitive 3D spatial models but also given access to advanced data analysis technologies. Its ability to represent multiple dimensions interactively addresses the challenges faced in traditional 2D visualization techniques. DataViz3D's integration with Looking Glass technology for holographic displays further augments its capability, offering immersive and detailed perspectives on data.

We hope DataViz3D will inspire more scholars to focus on dataset visualization and prove beneficial to peers in machine learning and deep learning, fostering advancements in these fields. The source code for DataViz3D is available for access and further development. It can be found at \href{https://github.com/shirou10086/FlexCAD-UCI-2021}{Github DataViz3D}, where users are encouraged to contribute to the project.
\clearpage
\bibliographystyle{elsarticle-harv} 
\bibliography{main}

\begin{thebibliography}{11}
\expandafter\ifx\csname natexlab\endcsname\relax\def\natexlab#1{#1}\fi
\providecommand{\url}[1]{\texttt{#1}}
\providecommand{\href}[2]{#2}
\providecommand{\path}[1]{#1}
\providecommand{\DOIprefix}{doi:}
\providecommand{\ArXivprefix}{arXiv:}
\providecommand{\URLprefix}{URL: }
\providecommand{\Pubmedprefix}{pmid:}
\providecommand{\doi}[1]{\href{http://dx.doi.org/#1}{\path{#1}}}
\providecommand{\Pubmed}[1]{\href{pmid:#1}{\path{#1}}}
\providecommand{\bibinfo}[2]{#2}
\ifx\xfnm\relax \def\xfnm[#1]{\unskip,\space#1}\fi
%Type = Inproceedings
\bibitem[{Barrett et~al.(2005)Barrett, Hunter, Miller, Hsu and Greenfield}]{Matplotlib}
\bibinfo{author}{Barrett, P.}, \bibinfo{author}{Hunter, J.}, \bibinfo{author}{Miller, J.}, \bibinfo{author}{Hsu, J.C.}, \bibinfo{author}{Greenfield, P.}, \bibinfo{year}{2005}.
\newblock \bibinfo{title}{matplotlib -- a portable python plotting package}.
%Type = Inproceedings
\bibitem[{Elewah et~al.(2021)Elewah, Badawi, Khalil, Rahnamayan and Elgazzar}]{3D-Radviz}
\bibinfo{author}{Elewah, A.}, \bibinfo{author}{Badawi, A.A.}, \bibinfo{author}{Khalil, H.}, \bibinfo{author}{Rahnamayan, S.}, \bibinfo{author}{Elgazzar, K.}, \bibinfo{year}{2021}.
\newblock \bibinfo{title}{3d-radviz: Three dimensional radial visualization for large-scale data visualization}, in: \bibinfo{booktitle}{2021 IEEE Congress on Evolutionary Computation (CEC)}, pp. \bibinfo{pages}{1037--1046}.
\newblock \DOIprefix\doi{10.1109/CEC45853.2021.9504983}.
%Type = Article
\bibitem[{Firat et~al.(2023)Firat, Swallow and Laramee}]{pcp-ed}
\bibinfo{author}{Firat, E.E.}, \bibinfo{author}{Swallow, B.}, \bibinfo{author}{Laramee, R.S.}, \bibinfo{year}{2023}.
\newblock \bibinfo{title}{Pcp-ed: Parallel coordinate plots for ensemble data}.
\newblock \bibinfo{journal}{Visual Informatics} \bibinfo{volume}{7}, \bibinfo{pages}{56--65}.
\newblock \URLprefix \url{https://www.sciencedirect.com/science/article/pii/S2468502X22001085}, \DOIprefix\doi{https://doi.org/10.1016/j.visinf.2022.10.003}.
%Type = Misc
\bibitem[{Fisher(1988)}]{misc_iris_53}
\bibinfo{author}{Fisher, R.A.}, \bibinfo{year}{1988}.
\newblock \bibinfo{title}{{Iris}}.
\newblock \bibinfo{howpublished}{UCI Machine Learning Repository}.
\newblock \bibinfo{note}{{DOI}: https://doi.org/10.24432/C56C76}.
%Type = Misc
\bibitem[{Gao et~al.(2019)Gao, Wang, Fan, Jiao, Xia, Xiong and Hong}]{holographic}
\bibinfo{author}{Gao, H.}, \bibinfo{author}{Wang, Y.}, \bibinfo{author}{Fan, X.}, \bibinfo{author}{Jiao, F.}, \bibinfo{author}{Xia, J.}, \bibinfo{author}{Xiong, W.}, \bibinfo{author}{Hong, M.}, \bibinfo{year}{2019}.
\newblock \bibinfo{title}{Dynamic high efficiency 3d meta-holography in visible range with large frames number and high frame rate based on space division multiplexing design}.
\newblock \href{http://arxiv.org/abs/1909.05642}{{\tt arXiv:1909.05642}}.
%Type = Misc
\bibitem[{Korting et~al.(2023)Korting, Lieberman, Matsudaira, Pei and Shen}]{Graphicalrep}
\bibinfo{author}{Korting, C.}, \bibinfo{author}{Lieberman, C.}, \bibinfo{author}{Matsudaira, J.}, \bibinfo{author}{Pei, Z.}, \bibinfo{author}{Shen, Y.}, \bibinfo{year}{2023}.
\newblock \bibinfo{title}{Visual inference and graphical representation in regression discontinuity designs}.
\newblock \href{http://arxiv.org/abs/2112.03096}{{\tt arXiv:2112.03096}}.
%Type = Article
\bibitem[{Li and Savvides(2013)}]{Irismethod}
\bibinfo{author}{Li, Y.H.}, \bibinfo{author}{Savvides, M.}, \bibinfo{year}{2013}.
\newblock \bibinfo{title}{An automatic iris occlusion estimation method based on high-dimensional density estimation}.
\newblock \bibinfo{journal}{IEEE Transactions on Pattern Analysis and Machine Intelligence} \bibinfo{volume}{35}, \bibinfo{pages}{784--796}.
\newblock \DOIprefix\doi{10.1109/TPAMI.2012.169}.
%Type = Misc
\bibitem[{LookingGlass(2024a)}]{holoplay}
\bibinfo{author}{LookingGlass}, \bibinfo{year}{2024}a.
\newblock \bibinfo{title}{Holoplay core js}.
\newblock \URLprefix \url{https://docs.lookingglassfactory.com/core/corejs}.
%Type = Misc
\bibitem[{LookingGlass(2024b)}]{LookingGlass_2024}
\bibinfo{author}{LookingGlass}, \bibinfo{year}{2024}b.
\newblock \bibinfo{title}{Immersive 3d for the masses. holograms and conversational ai characters.}
\newblock \URLprefix \url{https://lookingglassfactory.com/}.
%Type = Book
\bibitem[{Milner and Goodale(2006)}]{visualbrain}
\bibinfo{author}{Milner, D.}, \bibinfo{author}{Goodale, M.}, \bibinfo{year}{2006}.
\newblock \bibinfo{title}{{The Visual Brain in Action}}.
\newblock \bibinfo{publisher}{Oxford University Press}.
\newblock \URLprefix \url{https://doi.org/10.1093/acprof:oso/9780198524724.001.0001}, \DOIprefix\doi{10.1093/acprof:oso/9780198524724.001.0001}.
%Type = Misc
\bibitem[{Three.js(2024)}]{threejs}
\bibinfo{author}{Three.js}, \bibinfo{year}{2024}.
\newblock \URLprefix \url{https://threejs.org/docs/index.html#manual/en/introduction/Installation}.

\end{thebibliography}

\end{document}